\documentclass[letterpaper, 10 pt, journal, twoside]{ieeetran}

\usepackage[T1]{fontenc}

\usepackage{amsmath,amssymb,amsfonts}
\usepackage{bm}

\usepackage{graphicx}
\usepackage{wrapfig}
\usepackage{caption}
\usepackage{subcaption}
\usepackage{array}
\usepackage{booktabs}
\usepackage{multirow}

\usepackage[table,dvipsnames]{xcolor}

\usepackage{pifont}
\usepackage[nointegrals]{wasysym}
\usepackage{textcomp}

\usepackage[ruled,vlined]{algorithm2e}

\usepackage{soul}
\usepackage[normalem]{ulem}

\usepackage{mdframed}

\usepackage{cite}
\usepackage{hyperref}
\usepackage{cleveref}

\usepackage{nomencl}
\makenomenclature

\renewcommand{\hl}[1]{#1}

\begin{document}

\title{DB-VIO: Dual-Branch Visual Inertial Odometry with Enhanced Visual-Inertial Representation}

\author{anonymous authors}
\markboth{}%
{DB-VIO: Dual-Branch Visual-Inertial Odometry}
\author{Ziyu~Wan and Lin~Zhao
\thanks{The authors are with the Department of Electrical and Computer Engineering, National University of Singapore, Singapore. (e-mail: \tt \footnotesize {wziyu@u.nus.edu, elezhli@nus.edu.sg})%
}
}
\maketitle

\begin{abstract}
Visual inertial odometry (VIO) is essential for accurate 6-DoF motion estimation in mobile robotic systems. Recent learning-based VIO methods have shown promising progress, but they often rely on unified visual--inertial representations and a single temporal model for full-pose estimation, limiting their ability to capture the heterogeneous dynamics of rotation and translation. Moreover, monocular visual features often lack explicit geometric structure, while raw inertial encoding leaves the underlying rotational kinematics implicit, weakening the rotation-related cues in IMU features. To address these issues, we propose DB-VIO, a dual-branch visual inertial odometry framework with enhanced visual--inertial representation. DB-VIO incorporates depth cues to improve monocular visual perception, injects an explicit integrated-attitude prior to strengthen rotation-aware inertial representation, and decouples pose estimation into dedicated rotational and translational branches for motion-specific temporal modeling. Experiments on autonomous driving and aerial robot benchmarks show that DB-VIO achieves state-of-the-art performance, improving the corresponding baselines by 20\% on KITTI and 33\% on EuRoC. Notably, under the more agile motion patterns of EuRoC, DB-VIO improves the rotational metric by 65.7\% over prior methods. These results demonstrate the effectiveness and generalization of DB-VIO across different platforms and motion scenarios. 
\end{abstract}


\vspace{-0.3cm}

\section{Introduction}
\begin{figure}[!t]
  \centering
  \includegraphics[scale = 0.38]{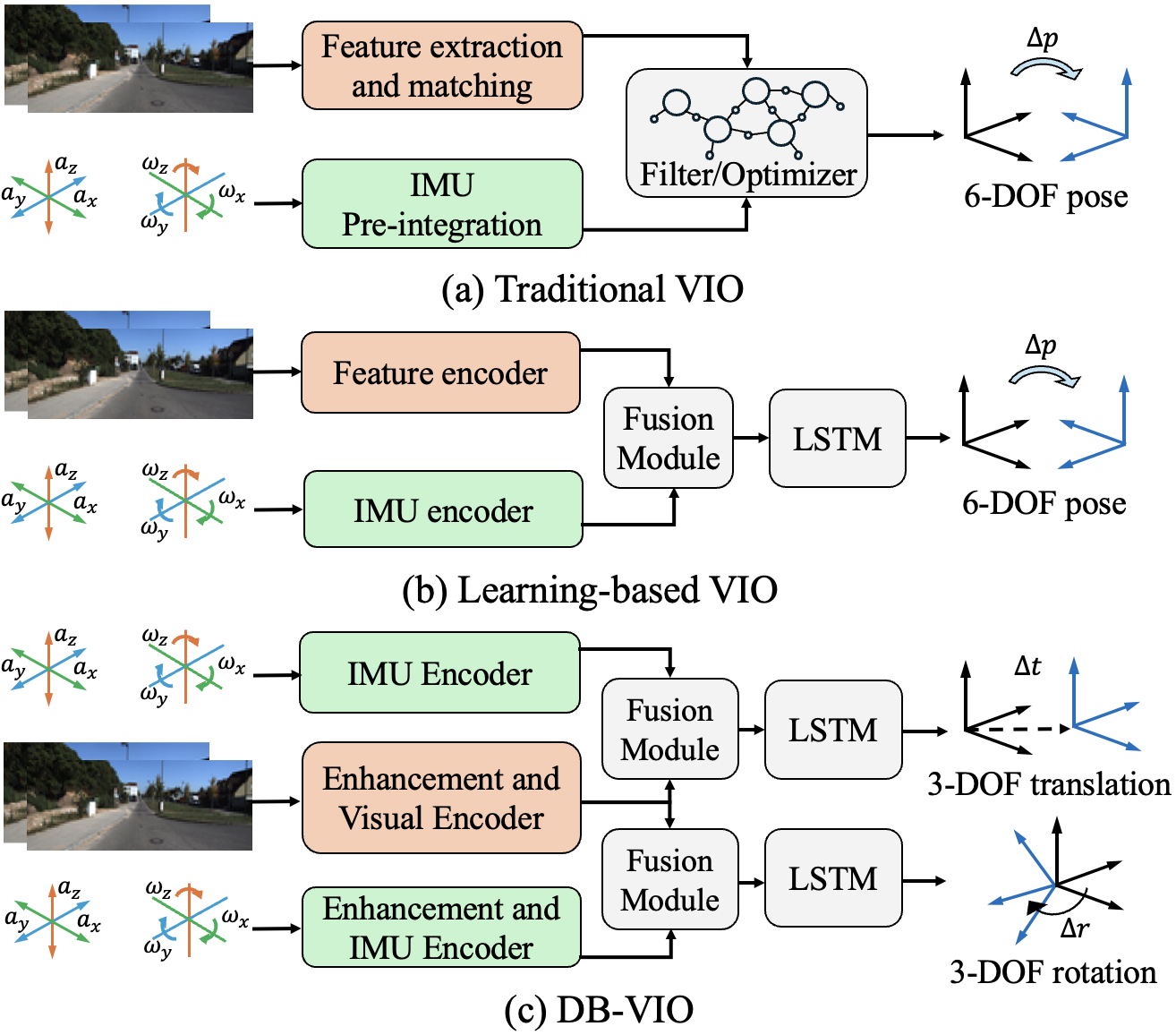}
    \caption{Conceptual overview of traditional VIO, learning-based VIO, and DB-VIO. Compared with existing deep VIO methods, DB-VIO strengthens visual-inertial representations and decouples temporal modeling into translation and rotation branches.}
    \vspace{-0.3cm}
  \label{fig:comparison}  
\end{figure}

\IEEEPARstart{R}{eliable} ego-motion estimation in GPS-denied environments benefits from both rich scene perception and continuous motion sensing. Cameras capture visual structure from the environment, while IMUs provide high-frequency inertial measurements \cite{scaramuzza2019visual}. By combining these complementary sensors, VIO has become an important solution for 6-DoF pose estimation in autonomous driving \cite{almalioglu2022selfvio}, robotic navigation \cite{tao2024learning}, and unmanned aerial vehicles \cite{qin2018vins}. 
Classical VIO systems usually rely on carefully designed visual front-ends, IMU preintegration, and nonlinear optimization to estimate camera motion, as illustrated in Fig.~\ref{fig:comparison}(a). 
Although effective in many scenarios, these methods remain sensitive to visual degradation caused by low texture and aggressive motion \cite{qin2018vins, campos2021orb, leutenegger2015keyframe, geneva2020openvins, lim2022uv, von2022dm}. Moreover, their performance and efficiency can be affected by accumulated IMU noise and bias, platform-dependent parameter tuning, unstable initialization, and the high computational cost of iterative optimization.

Recently, learning-based VIO methods have attracted increasing attention due to their ability to learn visual--inertial representations directly from data \cite{chen2019selective, han2019deepvio, yang2022efficient, chen2022learning}.
Compared with traditional geometric pipelines, deep VIO methods reduce the dependence on hand-crafted feature extraction and explicit optimization, and have shown promising robustness in complex environments \cite{qin2025ss, ma2026mixvio}. 
However, existing learning-based VIO frameworks still face several limitations.
Most methods rely on monocular RGB images as the primary visual input, whose learned representations are largely dominated by appearance cues such as texture, edges, and local semantics. Such representations do not explicitly capture geometry-aware information, including depth distribution, spatial relationships, and structural cues, which are crucial for accurate and robust ego-motion estimation.
In addition to visual representation, inertial encoding also remains insufficiently explored. 
Although IMU measurements provide high-frequency motion cues, existing fusion networks encode raw inertial sequences while leaving the underlying rotational kinematics implicit, forcing the network to approximate manifold-level integration and weakening the rotation-related cues in the learned IMU features. Meanwhile, long-horizon orientation is retained only implicitly in the recurrent hidden state, lacking an explicit attitude reference \cite{qiu2025airio}.
Furthermore, the temporal modeling strategy adopted by most deep VIO methods is still overly coupled. Most deep VIO methods employ a single recurrent module to regress the full 6-DoF pose, implicitly treating rotation and translation as a unified temporal prediction problem. In practice, rotational and translational motions may exhibit heterogeneous temporal dynamics and error propagation patterns, making a unified temporal model suboptimal.

To address these issues, we propose DB-VIO, a dual-branch visual inertial odometry framework with enhanced visual inertial representation. 
Instead of relying only on appearance-driven RGB features, DB-VIO augments monocular visual representation with relative depth cues estimated by Metric3D \cite{hu2024metric3d}, providing complementary geometric structure for visual--inertial feature learning.
For inertial representation, inspired by the observation that explicitly encoding attitude information benefits inertial feature learning \cite{qiu2025airio}, DB-VIO introduces an Attitude-Guided Encoding module that explicitly integrates gyroscope measurements into an attitude prior and injects it into IMU features, thereby strengthening rotation-aware inertial representation.
In addition to representation enhancement, DB-VIO further revisits the temporal modeling strategy for deep VIO. Rather than using a single recurrent estimator for the complete 6-DoF pose, DB-VIO decouples pose estimation into two dedicated branches for rotational and translational motion. This dual-branch design enables motion-specific temporal modeling and alleviates the interference between heterogeneous pose components.

We evaluate DB-VIO on both autonomous driving and aerial robot benchmarks, including the KITTI odometry dataset \cite{geiger2013vision} and the EuRoC MAV dataset \cite{burri2016euroc}. Experimental results show that DB-VIO achieves state-of-the-art performance across different platforms and motion scenarios. Specifically, DB-VIO improves the corresponding baselines by 20\% on KITTI and 33\% on EuRoC MAV, demonstrating the effectiveness of the proposed enhanced representation and dual-branch motion modeling.
The main contributions of this work are summarized as follows:
\begin{itemize}
    \item We propose DB-VIO, a dual-branch deep visual--inertial odometry framework that decouples rotational and translational motion modeling for more effective 6-DoF pose estimation.
    \item We design a dual representation enhancement strategy, including Depth-Guided Fusion (DGF) for visual features and Attitude-Guided Encoding (AGE) for inertial features. DGF enriches monocular RGB representation with Metric3D-estimated relative depth cues, while AGE integrates gyroscope measurements into an explicit attitude prior and enhances the IMU features.
    \item Extensive experiments on KITTI and EuRoC MAV demonstrate that DB-VIO achieves state-of-the-art performance and generalizes well across ground-vehicle and aerial-robot scenarios.
\end{itemize}

\section{Related Works}
\subsection{Traditional VIO Methods}
Traditional VIO methods estimate motion by exploiting geometric constraints between visual observations and inertial measurements. 
Filtering-based methods, represented by MSCKF \cite{mourikis2007multi} and its variants \cite{barrau2016invariant}, propagate IMU states and update them with visual constraints, providing efficient online estimation. Optimization-based methods, such as OKVIS \cite{leutenegger2015keyframe}, VINS-Mono \cite{qin2018vins}, and ORB-SLAM3 \cite{campos2021orb}, formulate VIO as a nonlinear optimization problem and jointly refine camera poses, IMU states, and visual landmarks using IMU preintegration and visual reprojection constraints. To improve robustness, later works further incorporate line features, point-line constraints, and more reliable initialization strategies \cite{von2022dm}.

Although these geometric pipelines have achieved strong performance, they rely on carefully designed visual front-ends, accurate initialization, and platform-dependent parameter tuning. Their accuracy may degrade when feature tracking becomes unreliable under low texture, motion blur, or aggressive motion. In addition, nonlinear optimization and keyframe management often introduce non-negligible computational cost, limiting their flexibility in resource-constrained or highly dynamic scenarios.
\subsection{Learning-based VIO Methods}
Learning-based VIO has been increasingly explored to reduce the dependence on hand-crafted geometric pipelines \cite{panadaptive}. VINet \cite{clark2017vinet} is one representative end-to-end framework that integrates visual and inertial features with an LSTM to estimate 6-DoF motion. Other works further combine neural networks with traditional estimation frameworks \cite{luo2025supervins, zuo2021codevio}, such as using learned features, learned depth, or bias correction modules within filtering or optimization pipeline, improving robustness while retaining geometric constraints. \cite{chen2019selective}

Recent studies have focused more on visual--inertial fusion and adaptive perception. Attention mechanisms, frequency module \cite{qin2025ss}, cross-modal interaction modules \cite{wang2024cmif}, and policy networks \cite{yang2022efficient, ma2026mixvio} have been introduced to improve feature selection, suppress unreliable measurements, or reduce unnecessary visual computation. These methods have improved the accuracy and efficiency of deep VIO under challenging sensing conditions. 
However, existing deep VIO methods usually insufficiently exploit the scene geometry and spatial layout implied by monocular images, as well as the rotational kinematics implicit in IMU measurements. Moreover, their unified temporal modeling of the full 6-DoF pose may obscure the heterogeneous dynamics of rotation and translation. 
\subsection{Enhanced Visual-Inertial Representation}
Effective visual--inertial representation is critical for learning-based VIO. 
For visual perception, monocular RGB features are mainly appearance-driven and may not explicitly encode scene geometry or spatial layout. Recent advances in monocular depth estimation, such as Metric3D \cite{hu2024metric3d} and Depth Anything \cite{yang2024depth}, have demonstrated strong zero-shot depth prediction ability through large-scale pretraining, making it possible to extract useful geometric cues from single RGB images. Existing work \cite{lai2025zerovo} has shown that depth-related cues can provide useful structural priors for visual odometry.
For inertial perception, the representation of IMU measurements also strongly affects motion estimation. Recent inertial odometry studies show that attitude-aware inertial modeling can improve IMU feature learning \cite{qiu2025airio}, especially under complex motion. 

\begin{figure*}[t]
  \centering
  \includegraphics[scale = 0.5]{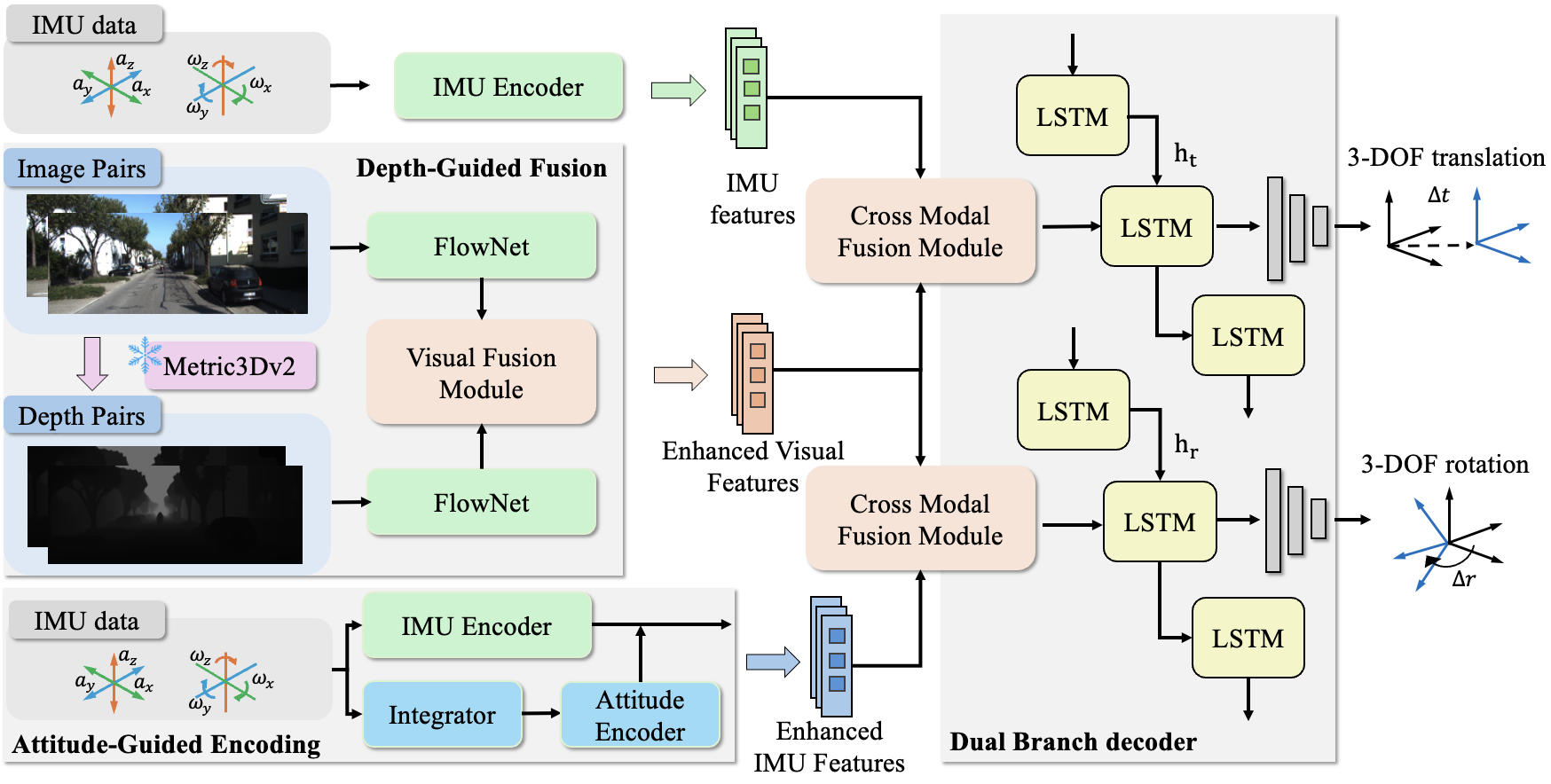}
    \caption{Overview of the proposed DB-VIO framework. DB-VIO enhances visual features via Depth-Guided Fusion and strengthens inertial features via Attitude-Guided Encoding. The resulting visual--inertial features are fused and processed by a dual-branch decoder, which separately models translation and rotation to estimate the final 6-DoF pose.}
    \vspace{-0.3cm}
  \label{fig:overview}  
\end{figure*}

Although prior studies have provided valuable insights into representation learning, how to effectively enhance visual--inertial representations for VIO remains insufficiently explored. 
To address this issue, DB-VIO introduces Depth-Guided Fusion (DGF) and Attitude-Guided Encoding (AGE), which enhance visual and inertial features using estimated depth cues and attitude integration, respectively.

\section{Method}
\subsection{Overall Architecture}
The overall architecture of the proposed DB-VIO is shown in Fig.~\ref{fig:overview}. The inputs are a sequence of monocular video frames $[F_i]_{i=1}^{N}$ and IMU measurements $[I_i]_{i=1}^{Nk}$ sampled at $k$ times the frame rate, together with an initial camera pose $P_0 \in \mathbf{SE}(3)$. Given the initial pose, the model estimates the camera poses $[P_i]_{i=1}^{N}$ for the entire trajectory, where $F_i \in \mathbb{R}^{n \times H \times W}$ denotes the video frame and $I_i \in \mathbb{R}^{6}$ contains the 3-axis accelerations and angular velocities. For each time index $t = 1, 2, \dots, N-1$, given two consecutive frames $F_{t \to t+1} = \{F_t, F_{t+1}\}$ and the IMU segment $I_{t \to t+1} = \{I_{tl}, \dots, I_{(t+1)l}\}$ collected between them, the model estimates the 6-DoF relative pose $T_{t \to t+1}$, which is decomposed into a rotational vector $\varphi_t \in \mathbb{R}^{3}$ and a translational vector $\phi_t \in \mathbb{R}^{3}$.

Unlike prior deep VIO methods that rely on appearance-driven visual features and a unified temporal model, DB-VIO is built around two ideas: enhancing visual--inertial representations, and decoupling the heterogeneous dynamics of rotation and translation. To this end, the framework comprises three stages. First, two representation-enhancement modules strengthen each modality. Depth-Guided Fusion (DGF) augments the monocular RGB input with relative depth cues estimated by Metric3Dv2~\cite{hu2024metric3d}, injecting geometric structure into the visual features $f_t^{v}$; 
Attitude-Guided Encoding (AGE) explicitly integrates the gyroscope measurements into an attitude prior and injects it into the inertial encoding to produce the enhanced inertial feature $f_t^{i}$, providing an explicit record of the orientation evolution that is otherwise left implicit in the raw IMU stream.
Second, the Cross Modal Fusion Module~\cite{qin2025ss} performs attention-based interaction between the enhanced visual and inertial features, producing complementary fused representations. Finally, rather than regressing the full 6-DoF pose with a single recurrent estimator, a dual-branch decoder processes the fused features through dedicated translational and rotational branches, each with its own recurrent module, enabling motion-specific temporal modeling and alleviating the interference between heterogeneous pose components. The estimated relative poses are accumulated from $P_0$ to recover the full trajectory.

\subsection{Depth-Guided Fusion}
Monocular RGB frames are dominated by appearance cues and lack explicit geometric structure, which limits the reliability of visual features for ego-motion estimation. To enrich the visual representation with geometry-aware information, we introduce Depth-Guided Fusion (DGF), which complements the RGB stream with relative depth cues.

Given two consecutive frames $F_{t \to t+1} = \{F_t, F_{t+1}\}$, we first estimate their relative depth maps with the pretrained Metric3D~\cite{hu2024metric3d} model, and stack the adjacent depth maps into a depth pair $D_{t \to t+1} = \{D_t, D_{t+1}\}$ aligned with the RGB pair. The RGB pair and the depth pair are then processed by two independent FlowNet-based encoders~\cite{dosovitskiy2015flownet}, which extract appearance-driven and geometry-aware motion features, respectively:
\begin{equation}
f_t^{rgb} = E_{rgb}(F_{t \to t+1}), \qquad f_t^{d} = E_{depth}(D_{t \to t+1}).
\end{equation}
The two feature streams are concatenated along the channel dimension and fused by the Visual Fusion Module, which applies convolutional layers to produce the enhanced visual feature:
\begin{equation}
f_t^{v} = \mathrm{VFM}\big([\,f_t^{rgb} \,\|\, f_t^{d}\,]\big),
\end{equation}
where $[\,\cdot \, \| \, \cdot\,]$ denotes channel-wise concatenation. By injecting depth-derived structural cues into the appearance features, DGF equips the visual representation $f_t^{v}$ with complementary geometric information, providing a more robust visual input for subsequent visual--inertial fusion.

\subsection{Attitude-Guided Encoding}
Unlike the accelerometer readings, the rotational motion between consecutive frames is determined by integrating the angular velocity, which is not a simple summation but a composition on the manifold $\mathrm{SO}(3)$. Given the gyroscope measurements $\{\omega_i\}$ within a time window, the relative rotation is obtained by chaining incremental rotations through the exponential map:
\begin{equation}
\Delta R = \prod_{i} \exp\!\big( (\omega_i)^{\wedge} \, \Delta t\big),
\label{eq:so3_integration}
\end{equation}
where $(\cdot)^{\wedge}: \mathbb{R}^3 \to \mathfrak{so}(3)$ maps a vector to its skew-symmetric matrix, $\exp(\cdot)$ denotes the exponential map from the Lie algebra $\mathfrak{so}(3)$ to the rotation group $\mathrm{SO}(3)$, and $\Delta t$ is the sampling interval. Because the composition in Eq.~\eqref{eq:so3_integration} is non-commutative, the mapping from raw angular velocities to relative rotation is inherently nonlinear and manifold-constrained. Requiring a recurrent or convolutional encoder to implicitly approximate this integration from raw IMU sequences is inefficient and difficult, which weakens the rotation-related cues in the learned inertial features.

To address this, Attitude-Guided Encoding (AGE) performs the rotational integration \emph{explicitly}. Concretely, given an initial attitude $R_0$ and the gyroscope measurements $\{\omega_i^{B}\}_{i=1}^{k}$ in the body frame, we integrate the attitude by chaining the incremental rotations by the extrinsic rotation $R_{bc}$:
\begin{equation}
R_k = R_0 \prod_{i=1}^{k} \exp\!\big( (R_{bc}\,\omega_i^{B})^{\wedge}\, \Delta t \big),
\label{eq:recursive_integration}
\end{equation}
and take the logarithmic map to obtain the corresponding rotation-vector representation:
\begin{equation}
\varphi_k = \log\!\big(R_k\big)^{\vee} \in \mathbb{R}^3,
\label{eq:log_map}
\end{equation}
where $\log(\cdot)$ is the inverse of the exponential map and $(\cdot)^{\vee}: \mathfrak{so}(3) \to \mathbb{R}^3$ is the inverse of $(\cdot)^{\wedge}$.
Applying Eq.~\eqref{eq:recursive_integration} and Eq.~\eqref{eq:log_map} at each IMU timestamp yields an explicit attitude sequence ${\varphi_k}$ that is temporally aligned with the raw inertial stream. The resulting sequence also explicitly preserves the evolution of orientation, providing historical attitude dynamics for subsequent temporal modeling.

We combine this sequence with the original IMU data to augment the inertial input, which is then processed by the inertial encoder:
\begin{equation}
f_t^{i} = E_{inertial}\big( I_{t \to t+1}, \varphi_{t \to t+1} \big),
\label{eq:age_encoding}
\end{equation}
where $\varphi_{t \to t+1}$ is the integrated attitude sequence over the corresponding window. The raw inertial segment without attitude augmentation is encoded as $\tilde{f}_t^{i} = E_{inertial}(I_{t \to t+1})$. In this way, AGE strengthens the rotation-aware information in $f_t^{i}$, relieving the encoder from learning the rotational dynamics from scratch.

\subsection{Dual-Branch Decoder}
DB-VIO decouples pose estimation into two dedicated branches, both branches share the depth-enhanced visual feature $f_t^{v}$ from DGF, while differing in their inertial input. The rotational branch is paired with the attitude-enhanced inertial feature $f_t^{i}$ from AGE, since rotational motion is most directly observed by the inertial modality, and its complex integration makes the explicit prior in $f_t^{i}$ particularly beneficial for this branch. The translational branch instead uses the original inertial feature $\tilde{f}_t^{i}$. Each branch employs its own Cross Fusion Module~\cite{qin2025ss} to perform attention-based cross-modal interaction between the shared visual feature and its respective inertial feature:
\begin{align}
f_t^{vi,r} &= \mathrm{CMF}_{r}\big(f_t^{v},\, f_t^{i}\big), \label{eq:cfm_rot}\\
f_t^{vi,t} &= \mathrm{CMF}_{t}\big(f_t^{v},\, \tilde{f}_t^{i}\big), \label{eq:cfm_trans}
\end{align}
where $f_t^{vi,r}, f_t^{vi,t}$ are the fused features for the rotational and translational branches, respectively.

Each fused feature is then processed by an independent LSTM that captures the motion-specific temporal dependencies of its branch, followed by a fully connected layer that regresses the corresponding pose component:
\begin{align}
\varphi_t &= \mathrm{FC}_{r}\big(\mathrm{LSTM}_{r}(f_t^{vi,r},\, h_{t-1}^{r})\big), \label{eq:rot_branch}\\
\phi_t &= \mathrm{FC}_{t}\big(\mathrm{LSTM}_{t}(f_t^{vi,t},\, h_{t-1}^{t})\big), \label{eq:trans_branch}
\end{align}
where $\varphi_t \in \mathbb{R}^3$ and $\phi_t \in \mathbb{R}^3$ are the estimated rotational and translational vectors, $h_{t-1}^{r}$ and $h_{t-1}^{t}$ denote the hidden states of the two LSTMs. By decoupling the two branches throughout fusion and temporal modeling, DB-VIO alleviates the interference between heterogeneous pose components. The estimated relative rotation and translation are assembled into the 6-DoF relative pose $T_{t \to t+1}$ and accumulated from $P_0$ to recover the full trajectory.

\subsection{Loss Function}
DB-VIO is trained with a combination of a per-frame pose loss and accumulated trajectory loss. The per-frame loss penalizes the relative pose error at each step. Denoting the estimated rotational and translational vectors as $\varphi_t$ and $\phi_t$ and their ground truth as $\varphi_t^{*}$ and $\phi_t^{*}$, the pose loss is a weighted sum of the rotational and translational errors:
\begin{equation}
\mathcal{L}_{pose} = \frac{1}{N}\sum_{t=1}^{N} \Big( \alpha \, \| \varphi_t - \varphi_t^{*} \|^2 + \| \phi_t - \phi_t^{*} \|^2 \Big),
\label{eq:pose_loss}
\end{equation}
where $\alpha$ balances the rotational and translational terms.

Since per-frame errors accumulate over time, we further supervise the accumulated trajectory to constrain long-horizon drift. The relative poses are integrated into global positions $\hat{p}_t$ and orientations $\hat{q}_t$, and compared against the ground-truth trajectory $p_t^{*}, q_t^{*}$. To make the supervision invariant to trajectory scale, each accumulated error is normalized by the corresponding path or rotation length:
\begin{equation}
\mathcal{L}_{acc}^{t} = \frac{\sum_{t} \| \hat{p}_t - p_t^{*} \|^2}{\sum_{t} \| \phi_t^{*} \|}, \qquad
\mathcal{L}_{acc}^{r} = \frac{\sum_{t} \| \hat{q}_t - q_t^{*} \|^2}{\sum_{t} \| \varphi_t^{*} \|},
\label{eq:accum_loss}
\end{equation}
where $\mathcal{L}_{acc}^{t}$ and $\mathcal{L}_{acc}^{r}$ denote the accumulated translational and rotational losses, normalized by the total translational and rotational path length, respectively.

The overall training objective combines the three terms:
\begin{equation}
\mathcal{L} = \mathcal{L}_{pose} + \lambda_1 \, \mathcal{L}_{acc}^{t} + \lambda_2 \, \mathcal{L}_{acc}^{r},
\label{eq:total_loss}
\end{equation}
where $\lambda_1$ and $\lambda_2$ balance the per-frame and accumulated supervision.

\section{EXPERIMENTS} \label{sec:exp}
We conduct extensive experiments to answer three questions that motivate the design of DB-VIO:
(1) Do rotational and translational motions exhibit heterogeneous temporal characteristics that justify modeling them with separate branches?
(2) Does DB-VIO generalize across platforms with distinct motion patterns?
(3) How does each proposed component contribute to the final performance?

To answer (1), we perform a spectral analysis of the ground-truth motion dynamics, examining the power spectral density and temporal autocorrelation of rotational and translational motion. To answer (2), we evaluate DB-VIO on the KITTI odometry dataset~\cite{geiger2013vision} for autonomous driving and the EuRoC MAV dataset~\cite{burri2016euroc} for aerial navigation, and compare against both traditional and learning-based VIO methods. To answer (3), we conduct ablation studies on the Depth-Guided Fusion, Attitude-Guided Encoding, and dual-branch decoder. All models are trained on a workstation with an AMD 7965WX CPU and an NVIDIA RTX 4090 GPU, and evaluated on an Intel Core i7-14700KF CPU with an NVIDIA RTX 4080 GPU.

\subsection{Motion Dynamics Analysis}
The dual-branch design of DB-VIO rests on the premise that rotational and translational motions possess different temporal characteristics. To verify this premise directly from data, we analyze the ground-truth motion dynamics of both datasets. For each sequence, we take the angular-velocity and linear-velocity time series as the rotational and translational dynamics, and normalize each axis to zero mean and unit variance so that the comparison reflects temporal structure rather than the raw $\mathrm{rad/s}$ versus $\mathrm{m/s}$ magnitude. We then characterize them from two complementary views: the power spectral density (PSD), computed with Welch's method, describes how motion energy is distributed over frequency, while the autocorrelation function (ACF) describes how quickly the motion decorrelates over time.

\begin{figure}[!t]
  \centering
  \includegraphics[scale = 0.48]{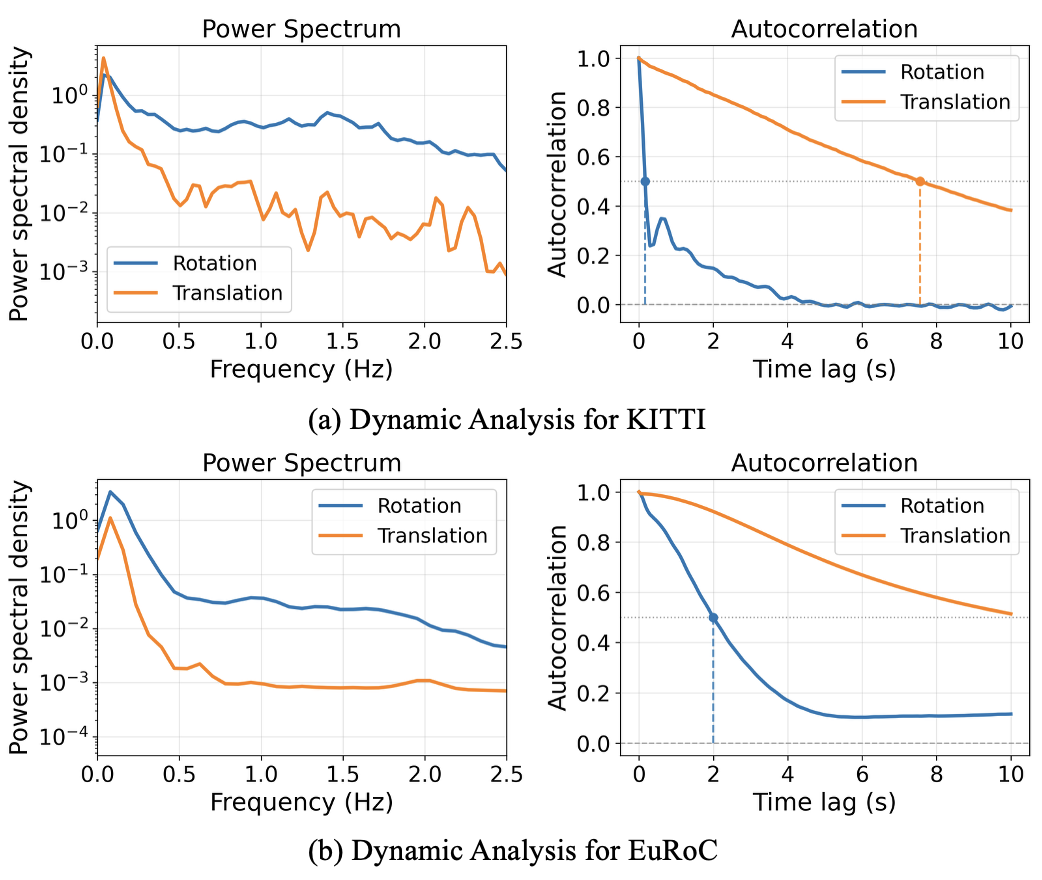}
    \caption{Model-free analysis of ground-truth motion dynamics on KITTI (a) and EuRoC (b). For each dataset, the left panel shows the power spectral density and the right panel the temporal autocorrelation of rotational (blue) and translational (orange) motion.}
    \label{fig:motion_analysis}
\end{figure}

\noindent\textbf{Frequency-domain structure.}
As shown in the left panels of Fig.~\ref{fig:motion_analysis}, translational motion concentrates its energy at low frequencies on both datasets. On KITTI, the two components are clearly separated: only $42.2\%$ of the rotational energy lies below $0.5\,\mathrm{Hz}$, compared with $91.4\%$ for translation, indicating that rotation carries a much larger high-frequency component. 

\noindent\textbf{Temporal correlation.}
The autocorrelation curves in the right panels of Fig.~\ref{fig:motion_analysis} expose an even sharper and more consistent contrast. On KITTI, the rotational autocorrelation drops below $0.5$ within only $0.17\,\mathrm{s}$, while translation stays correlated for about $7.57\,\mathrm{s}$. On EuRoC, rotation decorrelates within about $2.0\,\mathrm{s}$, whereas translation remains above $0.5$ throughout the entire $10\,\mathrm{s}$ window. On both a ground vehicle and an aerial robot, rotation thus loses temporal coherence far faster than translation, revealing fundamentally different memory lengths for the two motion components.

\noindent\textbf{Conclusion.}
Across two platforms with very different motion profiles, a near-planar ground vehicle (KITTI) and an agile aerial robot performing six-DoF maneuvers (EuRoC), rotational and translational motions consistently exhibit different temporal dynamics, with rotation being a shorter-memory signal than translation. Such heterogeneity makes a single shared recurrent model ill-suited to capture both simultaneously, and directly motivates the dual-branch design that models rotation and translation with dedicated temporal modules.

\subsection{KITTI Odometry}
\begin{table*}[!t]
\centering
\caption{Quantitative Results on the KITTI}
\label{tab:kitti}
\setlength{\tabcolsep}{3pt}          
\fontsize{7.5pt}{9pt}\selectfont     
\begin{tabular}{lll cc cc cc cc}
\toprule
\multirow{2}{*}{\textbf{Method}} & \multirow{2}{*}{\textbf{Reference}} & \multirow{2}{*}{\textbf{Mode}} & \multicolumn{2}{c}{\textbf{Seq. 05}} & \multicolumn{2}{c}{\textbf{Seq. 07}} & \multicolumn{2}{c}{\textbf{Seq. 10}} & \multicolumn{2}{c}{\textbf{Ave.}} \\
\cmidrule(lr){4-5} \cmidrule(lr){6-7} \cmidrule(lr){8-9} \cmidrule(lr){10-11}
 & & & $t_{rel}(\%)\downarrow$ & $r_{rel}(^{\circ}/\text{km})\downarrow$ & $t_{rel}(\%)\downarrow$ & $r_{rel}(^{\circ}/\text{km})\downarrow$ & $t_{rel}(\%)\downarrow$ & $r_{rel}(^{\circ}/\text{km})\downarrow$ & $t_{rel}(\%)\downarrow$ & $r_{rel}(^{\circ}/\text{km})\downarrow$ \\
\midrule
ORB-SLAM2 \cite{mur2017orb} & TRO 2017 & Geo. & \textbf{9.12} & \textbf{2.0} & 10.34 & \textbf{3.0} & \textbf{4.04} & \textbf{3.0} & \textbf{7.83} & \textbf{2.7} \\
VINS-Mono \cite{qin2018vins} & TRO 2018 & Geo. & 11.6 & 12.6 & \textbf{10.0} & 17.2 & 16.5 & 23.4 & 12.7 & 17.7 \\
\midrule
VIOLearner \cite{shamwell2018vision} & IROS 2018 & Self-Sup. & 3.00 & \textbf{14.0} & 3.60 & 20.6 & 2.04 & 13.7 & 2.88 & \textbf{16.1} \\
DeepVIO \cite{han2019deepvio} & IROS 2019 & Self-Sup. & \textbf{2.86} & 23.2 & \textbf{2.71} & \textbf{16.6} & \textbf{0.85} & \textbf{10.3} & \textbf{2.14} & 16.7 \\
\midrule
ATVIO \cite{liu2021atvio} & ICASSP 2021 & Sup. & 4.93 & 24.0 & 3.78 & 25.9 & 5.71 & 29.6 & 4.81 & 26.5 \\
Soft Fusion-VIO \cite{chen2022learning} & TNNLS 2025 & Sup. & 4.65 & 18.3 & 4.36 & 21.9 & 8.35 & 20.1 & 5.78 & 20.1 \\
Hard Fusion-VIO \cite{chen2022learning} & TNNLS 2025 & Sup. & 4.25 & 16.7 & 4.46 & 21.7 & 5.81 & 15.5 & 4.84 & 18.0 \\
VS-VIO \cite{yang2022efficient} & ECCV 2022 & Sup. & 2.61 & 10.6 & 1.83 & 13.5 & 3.11 & 11.2 & 2.52 & 11.8 \\
Gravity-Shift-VIO \cite{chen2023gravity} & IJCNN 2023 & Sup. & 2.98 & 14.0 & 2.76 & 18.7 & 4.45 & 12.1 & 3.39 & 14.9 \\
CAP-VIO \cite{pajula2023novel} & JISPIN 2024 & Sup. & 4.76 & 13.9 & 5.18 & 25.5 & 6.55 & 18.3 & 5.49 & 19.2 \\
CMIF-VIO \cite{wang2024cmif} & RA-L 2025 & Sup. & 2.65 & 9.5 & 1.76 & 9.8 & 3.45 & 10.4 & 2.62 & 9.9 \\
SS-VIO \cite{qin2025ss} & RA-L 2026 & Sup. & 2.43 & 9.2 & 2.03 & 8.6 & 2.24 & 9.2 & 2.23 & 9.0 \\
\midrule
\textbf{DB-VIO (Ours)} & & Sup. & \textbf{1.86} & \textbf{7.2} & \textbf{1.50} & \textbf{6.1} & \textbf{1.96} & \textbf{7.3} & \textbf{1.77} & \textbf{6.9} \\
\bottomrule
\end{tabular}
\end{table*}

\noindent\textbf{Dataset.} 
The KITTI odometry dataset provides 22 driving sequences spanning urban, rural, and highway scenarios, and is a standard benchmark for autonomous-driving VIO. Following the common supervised split~\cite{qin2025ss, yang2022efficient}, we train on sequences 00, 01, 02, 04, 06, 08, and 09, and evaluate on sequences 05, 07, and 10. Sequence 03 is dropped for lacking IMU data, and sequences 11--22 are unusable as they provide no ground-truth poses. We use a monocular setup with the left-camera images at $10\,\mathrm{Hz}$, resized to $512 \times 256$. Since the $100\,\mathrm{Hz}$ IMU stream is not strictly synchronized with the frames, we interpolate it to the image timestamps, associating $11$ consecutive six-axis IMU samples with each image pair. For evaluation, we follow the standard KITTI odometry protocol and report the average relative translation error $t_{\mathrm{rel}}$ and relative rotation error $r_{\mathrm{rel}}$ over all subsequences of lengths from $100$ to $800\,\mathrm{m}$. Here, $t_{\mathrm{rel}}$ is measured in percentage, and $r_{\mathrm{rel}}$ is measured in degrees per kilometer.

\noindent\textbf{Implementation.}
We train the model with a batch size of 16. The training process consists of two stages. First, we employ a single-stream LSTM to train the visual frontend, enabling it to learn sufficiently informative representations for both rotation and translation estimation. After obtaining the trained visual frontend, we freeze its parameters and subsequently train the dual-stream LSTM and the AGE module. We set the loss-balancing coefficient $\alpha$ to 100. 

\noindent\textbf{Results.}
Table~\ref{tab:kitti} reports the quantitative comparison between DB-VIO and representative geometry-based, self-supervised, and supervised VIO methods. DB-VIO achieves translation errors of $1.86\%$, $1.50\%$, and $1.96\%$ on sequences 05, 07, and 10, respectively, while the corresponding rotation errors are $7.2^{\circ}/\mathrm{km}$, $6.1^{\circ}/\mathrm{km}$, and $7.3^{\circ}/\mathrm{km}$. Averaged over the three test sequences, our method obtains a translation error of $1.77\%$ and a rotation error of $6.9^{\circ}/\mathrm{km}$, achieving the best overall performance among all evaluated learning-based methods. In particular, DB-VIO consistently maintains low errors across the three sequences.

\begin{figure}[!t]
  \centering
  \includegraphics[scale = 0.4]{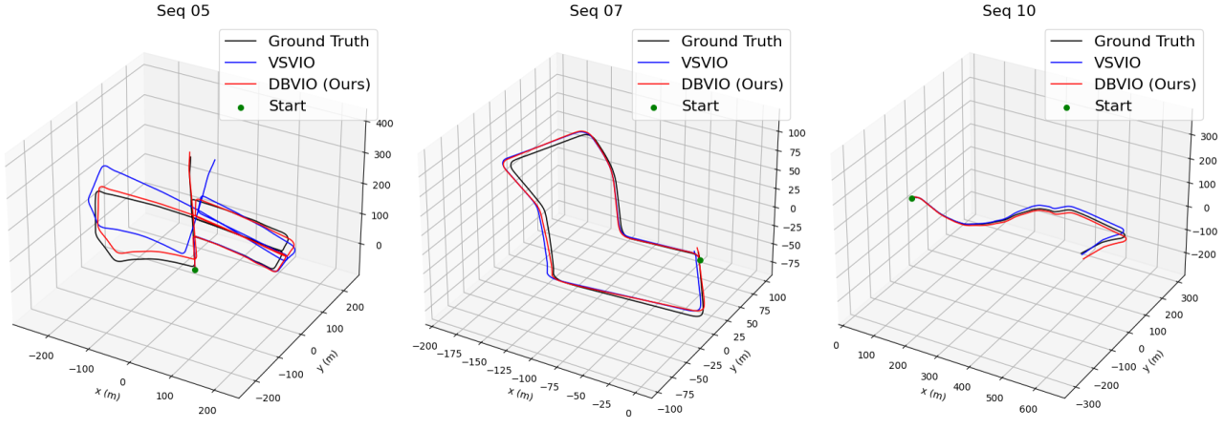}
    \caption{Qualitative trajectory comparison on test KITTI sequences.}
    \label{fig:kitti_traj}
\end{figure}

Compared with geometry-based approaches, DB-VIO achieves substantially better translation accuracy, while ORB-SLAM2 retains an advantage in rotation estimation due to its explicit geometric optimization.
DB-VIO also consistently outperforms both self-supervised and supervised learning-based methods. In particular, compared with SS-VIO, the strongest supervised baseline, our method reduces the average translation and rotation errors by $20.6\%$ and $23.3\%$, respectively. The improvements are consistently observed across all three test sequences, demonstrating that DB-VIO generalizes well across diverse driving scenarios. Fig.~\ref{fig:kitti_traj} provides a qualitative comparison of the estimated trajectories on test KITTI sequences. Compared with VSVIO, DB-VIO produces trajectories that align more closely with the ground truth, with smaller accumulated drift and more accurate motion estimation.
The superior performance on KITTI suggests that DB-VIO can exploit complementary visual and inertial cues more effectively than existing learning-based methods. 
By enhancing monocular visual representation, strengthening rotation-related inertial features, and adopting a dual-branch design, DB-VIO achieves more accurate and balanced 6-DoF pose estimation.
The consistent improvements across all test sequences further indicate that DB-VIO maintains stable performance under different outdoor driving conditions. Moreover, DB-VIO achieves an end-to-end inference frequency of 12.6 Hz, exceeding KITTI’s 10 Hz image acquisition rate, while its average forward-pass latency is 63 ms, demonstrating its real-time capability and practical applicability.

\begin{table}
\centering
\caption{Performance Under Different Settings}
\label{tab:ablation}
\renewcommand{\arraystretch}{1.2}
\begin{tabular}{cccccc}
\hline
\multirow{2}{*}{DB} & \multirow{2}{*}{DGF} & \multirow{2}{*}{AGE} & \multicolumn{2}{c}{Ave.} \\
\cline{4-5}
 & & & $t_{rel}(\%)\downarrow$ & $r_{rel}(^\circ/\text{km})\downarrow$  \\
\hline
\ding{55} & \ding{55} & \ding{55} & 2.18 & 9.2  \\
\ding{55} & \ding{51} & \ding{55} & 1.92 & 8.6  \\
\ding{51} & \ding{51} & \ding{55} & 1.85 & 8.3  \\
\ding{55} & \ding{51} & \ding{51} & 2.01 & 7.8  \\
\ding{51} & \ding{51} & \ding{51} & \textbf{1.77} & \textbf{6.9} \\
\hline
\end{tabular}
\end{table}

\noindent\textbf{Ablation studies.}
To evaluate the effectiveness of the proposed components, we conduct an ablation study on the KITTI benchmark.  As shown in Table~\ref{tab:ablation}, we progressively enable the dual-branch decoder (DB), Depth-Guided Fusion (DGF), and Attitude-Gated Encoding (AGE), while keeping the training and evaluation settings unchanged. 

Introducing DGF clearly improves translation accuracy, indicating that depth-guided visual representation provides useful geometric cues for motion estimation. 
The dual-branch decoder and AGE jointly improve rotation estimation: the dual-branch design reduces the coupling between translational and rotational temporal dynamics, while AGE strengthens rotation-related inertial features through attitude-aware encoding. 
Their combination leads to a more effective rotation representation and more accurate pose prediction.

With all components enabled, DB-VIO achieves the best overall performance in both translation and rotation. 
The results show that DGF, AGE, and the dual-branch decoder provide complementary benefits for representation enhancement and motion-specific temporal modeling.

\subsection{EuRoC MAV}
\noindent\textbf{Dataset.}
The EuRoC MAV dataset is a widely used benchmark for evaluating visual--inertial odometry on aerial robots. 
It contains synchronized stereo images, IMU measurements, and ground-truth trajectories collected by a micro aerial vehicle in indoor environments with different motion speeds and scene difficulties. 
Following~\cite{qin2025ss}, we use MH\_03\_medium, V1\_03\_difficult, and V2\_03\_difficult for testing, and use the remaining sequences for training. 
We adopt a monocular setting using the left-camera images, which are captured at $20\,\mathrm{Hz}$, while the IMU measurements are recorded at $200\,\mathrm{Hz}$. 
The image and IMU streams are temporally aligned through linear interpolation. 
For evaluation, we report the absolute trajectory error (ATE), which measures the global consistency between the estimated trajectory and the ground truth. 
Following~\cite{qin2018vins}, we also evaluate the average relative translation error $t_{\mathrm{rel}}$ and relative rotation error $r_{\mathrm{rel}}$. 
Considering that EuRoC trajectories are shorter than KITTI driving sequences, the relative errors are computed over segment lengths of $[20,30,40,50,60,70,80,90,100,110]\,\mathrm{m}$.

\noindent\textbf{Implementation.}
We train DB-VIO on EuRoC with a batch size of 16 using a three-stage training strategy. 
Following the KITTI setting, the visual frontend is first trained to learn informative representations for pose estimation. 
Afterward, we freeze the visual frontend and sequentially train the dual-branch temporal decoder and the AGE module. 
The loss-balancing coefficient $\alpha$ is set to 100. 
For the accumulated loss, we set $\lambda_1=0.2$ and $\lambda_2=16$.

\noindent\textbf{Results.}
Table~\ref{tab:euroc} presents the quantitative results on the EuRoC MAV dataset. 
Since EuRoC provides images at $20\,\mathrm{Hz}$, introducing depth estimation at every frame may increase the computational burden for online inference. 
Therefore, we take the variant without depth input as the main DB-VIO result, while also reporting the full DB-VIO with depth cues for completeness. The main DB-VIO variant achieves an inference frequency of 202.3 Hz, while the full DB-VIO with depth cues runs at 14.0 Hz due to the additional depth-estimation overhead.

Compared with existing learning-based VIO methods, DB-VIO achieves the best overall performance. 
Without using depth cues, DB-VIO improves over the previous state-of-the-art SS-VIO by $2.3\%$ in relative translation error, $65.2\%$ in relative rotation error, and $33.5\%$ in ATE. 
The large reduction in rotation error demonstrates the effectiveness of the proposed attitude-gated inertial encoding and dual-branch temporal modeling under agile UAV motion. 
When depth cues are further introduced, DB-VIO achieves additional improvements in translation accuracy and ATE, reducing the relative translation error and ATE by $5.1\%$ and $37.4\%$ compared with SS-VIO, respectively. 
This indicates that depth-guided visual representation provides useful geometric support for trajectory estimation.
Although geometry-based methods such as ORB-SLAM3 and DM-VIO achieve good performance on EuRoC, their performance relies heavily on successful initialization and carefully engineered geometric pipelines. In our implementation of these baselines, we also observed initialization failures, and prior results indicate that such methods may fail to converge when transferred to KITTI or other robotic platforms, revealing their limited robustness across different sensing and motion conditions \cite{qin2025ss}.
\begin{table}[t]
\centering
\caption{Quantitative Results on the EuRoC}
\label{tab:euroc}
\setlength{\tabcolsep}{3.5pt}
\begin{tabular}{l c c c c}
\toprule
\multirow{2}{*}{\textbf{Method}} & \multirow{2}{*}{\textbf{Mode}} & \multicolumn{3}{c}{\textbf{Ave.}} \\
\cmidrule(lr){3-5}
 & & $t_{\mathrm{rel}}(\%)\downarrow$ 
 & $r_{\mathrm{rel}}(^{\circ}/\mathrm{km})\downarrow$ 
 & $\mathrm{ATE}(\mathrm{m})\downarrow$ \\
\midrule
ORB-SLAM3~\cite{campos2021orb} & Geo. & \textbf{0.32} & \textbf{31.91} & \textbf{0.06} \\
DM-VIO~\cite{von2022dm} & Geo. & 0.99 & 87.10 & 0.37 \\
\midrule
Soft Fusion-VIO~\cite{chen2022learning} & Sup. & 17.68 & 3743.42 & 6.04 \\
Hard Fusion-VIO~\cite{chen2022learning} & Sup. & 14.43 & 3679.97 & 8.85 \\
VS-VIO~\cite{yang2022efficient} & Sup. & 6.30 & 356.77 & 2.92 \\
SS-VIO \cite{qin2025ss} & Sup. & 5.29 & 185.56 & 2.03 \\
DB-VIO (w. depth) & Sup. & \textbf{5.02} & 66.21 & \textbf{1.27} \\
DB-VIO & Sup. & 5.17 & \textbf{64.55} & 1.35 \\
\bottomrule
\end{tabular}
\end{table}

\begin{table}
\centering
\caption{Performance Under Different Settings}
\label{tab:euroc_ablation}
\renewcommand{\arraystretch}{1.2}
\begin{tabular}{ccccccc}
\hline
\multirow{2}{*}{DB} & \multirow{2}{*}{DGF} & \multirow{2}{*}{AGE} & \multicolumn{3}{c}{Ave.} \\
\cline{4-6}
 & & & $t_{rel}(\%)\downarrow$ & $r_{rel}(^\circ/\text{km})\downarrow$ & $ATE\downarrow$ \\
\hline
\ding{55} & \ding{55} & \ding{55} & 6.60 & 147.46 & 1.73 \\
\ding{51} & \ding{55} & \ding{55} & 5.18 & 70.48 & 1.36\\
\ding{51} & \ding{55} & \ding{51} & 5.17 & \textbf{64.55} & 1.35\\
\ding{55} & \ding{55} & \ding{51} & 6.43 & 84.47 & 1.69\\
\ding{51} & \ding{51} & \ding{51} & \textbf{5.02} & 66.21 & \textbf{1.27} & \\
\hline 
\end{tabular}
\end{table}

\noindent\textbf{Ablation Study.}
Table~\ref{tab:euroc_ablation} reports the ablation results on EuRoC. 
Different from KITTI, EuRoC contains more agile UAV motions with frequent rotation changes and more complex motion dynamics. 
Under this setting, the dual-branch decoder shows a particularly strong contribution. 
Introducing DB alone significantly reduces both translation and rotation errors, indicating that decoupling translational and rotational temporal modeling is especially beneficial for aerial motion estimation.
AGE further improves rotation accuracy by strengthening rotation-related inertial representation. 
When combined with the dual-branch decoder, AGE achieves the lowest rotation error, showing that attitude-aware inertial encoding is complementary to motion-specific temporal modeling. 
DGF mainly improves translation and trajectory accuracy, as the depth-guided visual cues provide additional geometric structure for estimating motion scale and reducing trajectory drift.
With all components enabled, DB-VIO achieves the best overall performance in translation error and ATE, while maintaining competitive rotation accuracy. 

\section{\texorpdfstring{\hl{CONCLUSIONS}}{CONCLUSIONS}}
In this paper, we presented DB-VIO, a dual-branch deep visual--inertial odometry framework with enhanced visual--inertial representation. 
DB-VIO improves monocular visual representation through Depth-Guided Fusion, which introduces estimated depth cues to provide additional geometric structure. 
For inertial representation, Attitude-Gated Encoding adaptively strengthens rotation-related IMU features. 
In addition, DB-VIO decouples temporal pose modeling into dedicated translation and rotation branches, enabling more effective modeling of heterogeneous motion dynamics.

Experiments on KITTI and EuRoC MAV demonstrate that DB-VIO achieves state-of-the-art performance among learning-based VIO methods across both ground-vehicle and aerial-robot scenarios. Ablation studies further confirm the complementary effects of Depth-Guided Fusion, Attitude-Gated Encoding, and the dual-branch temporal decoder. 
In future work, we will further explore more efficient representation learning and extend DB-VIO to more challenging real-world environments.

\bibliographystyle{IEEEtran}
\bibliography{IEEEabrv,main}

\vfill

\end{document}